%% file: main.tex
\documentclass[10pt,twocolumn,letterpaper]{article}
\usepackage[accsupp]{axessibility}
\usepackage{cvpr}              
\input{preamble}
\definecolor{cvprblue}{rgb}{0.21,0.49,0.74}
\usepackage[pagebackref,breaklinks,colorlinks,allcolors=cvprblue]{hyperref}
\usepackage{algorithm}
\usepackage{algorithmic}
\usepackage{pifont}
\usepackage{booktabs}
\usepackage{amsmath}
\usepackage{newfloat}
\usepackage{listings}
\def\paperID{36476} 
\def\confName{CVPR}
\def\confYear{2026}

\title{Dehallu3D: Hallucination-Mitigated 3D Generation from Single Image via Cyclic View Consistency Refinement}

\author{
    Xiwen Wang \quad 
    Shichao Zhang \quad 
    Ruowei Wang \quad 
    Mao Li \quad 
    Chenyu Zhou \\
    Ji-Zhe Zhou\thanks{Corresponding authors.} \quad 
    Qijun Zhao\footnotemark[1] \quad 
    Hailun Zhang \\
    Sichuan University, China \\
}

\begin{document}
\maketitle
\input{sec/0_abstract}    
\input{sec/1_intro}
\input{sec/2_relatedwork}

\input{sec/4_method}
\input{sec/5_exp}

\input{sec/6_conclusion}
\input{sec/acknowledgement}

{
    \small
    \bibliographystyle{ieeenat_fullname}
    \bibliography{main}
}



\end{document}

%% file: sec/0_abstract.tex
\begin{abstract}
Large 3D reconstruction models have revolutionized the 3D content generation field, enabling broad applications in virtual reality and gaming.
Just like other large models, large 3D reconstruction models suffer from hallucinations as well, introducing structural outliers (e.g., odd holes or protrusions) that deviate from the input data. However, unlike other large models, hallucinations in large 3D reconstruction models remain severely underexplored, leading to malformed 3D-printed objects or insufficient immersion in virtual scenes.
Such hallucinations majorly originate from that existing methods reconstruct 3D content from sparsely generated multi-view images which suffer from large viewpoint gaps and discontinuities. 
To mitigate hallucinations by eliminating the outliers, we propose \textbf{Dehallu3D} for 3D mesh generation.
Our key idea is to design a balanced multi-view continuity constraint to enforce smooth transitions across dense intermediate viewpoints, while avoiding over-smoothing that could erase sharp geometric features.
Therefore, Dehallu3D employs a plug-and-play optimization module with two key constraints: (i) adjacent consistency to ensure geometric continuity across views, and (ii) adaptive smoothness to retain fine details.
We further propose the Outlier Risk Measure (ORM) metric to quantify geometric fidelity in 3D generation from the perspective of outliers. Extensive experiments show that Dehallu3D achieves high-fidelity 3D generation by effectively preserving structural details while removing hallucinated outliers.
\end{abstract}

%% file: sec/1_intro.tex
\section{Introduction}
\label{sec:intro}
Generating 3D contents from 2D images is crucial for applications ranging from augmented reality and virtual reality, to 3D printing.
Despite significant advancements~\cite{li2024icontrol3d, li2024loopgaussian, song2024magiccartoon, pan2024harmonicnerf, zheng2024sketch3d, genudc}, the process of generating accurate 3D content from a single image continues to pose substantial challenges. 

Diffusion models \cite{saharia2022photorealistic} have achieved profound milestones in 2D generation, paving the way for 3D synthesis. DreamFusion and subsequent work \cite{poole2022dreamfusion, shen2023make, liu2024sherpa3d} introduced Score Distillation Sampling (SDS) to adapt pretrained 2D diffusion models for 3D object generation. However, SDS-based methods suffer from issues like the Janus problem \cite{poole2022dreamfusion} due to the absence of robust 3D priors. To address this, transformer-based Large Reconstruction Models (LRMs) \cite{hong2023lrm, xu2024instantmesh, lu2024large, tochilkin2024triposr} integrate 3D data for feedforward generation, but still face challenges in achieving accurate reconstruction.
As a solution to the above issues, a common paradigm trains 2D diffusion models to generate multi-view images for 3D reconstruction via neural rendering or multi-view stereo. \cite{long2024wonder3d, wu2024unique3d, yang2024hi3d, yu2025fancy123, tang2025cycle3d}.

\begin{figure}[t]
    \centering
    \includegraphics[width=1\linewidth]{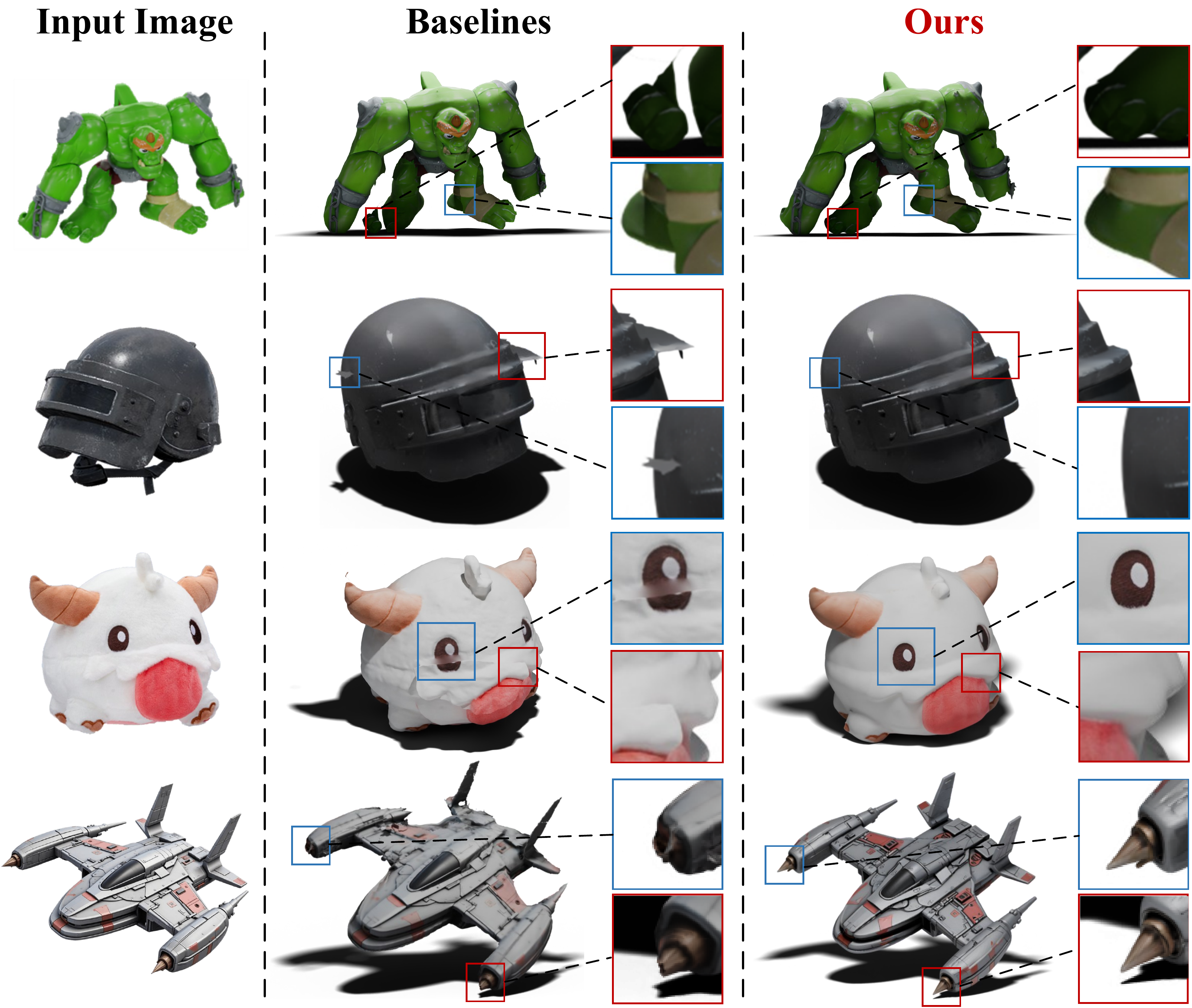}
    \caption{Our \textbf{Dehallu3D} generates high-quality and high-fidelity 3D meshes, effectively mitigating mesh outliers while adaptively preserving sharp features. The red and blue boxes highlight noticeable outlier regions in the Baseline method, contrasted with the corresponding regions in Dehallu3D (Ours) to showcase improvements.}
    \label{fig_title}
\end{figure}

These large 3D reconstruction models, similar to other large models, frequently encounter hallucinations. 
Such hallucinations typically manifest as outputs that diverge from ground truth, as shown in Figure~\ref{fig_title}, where they give rise to outliers in reconstructed contents.
However, in contrast to other large models, the hallucinations observed in 3D reconstruction models remain underexplored by current research. Consequently, these outliers pose significant challenges for downstream applications. In 3D printing, even minor geometric deviations can lead to manufacturing failures. Similarly, in gaming experiences, visual inconsistencies introduced by such outliers severely disrupt user immersion. 
We argue that such hallucinations originates from the inconsistent sparse multi-view images, which suffer from large viewpoint gaps and poor continuity. This misleads the model into introducing intricate details that are not supported by the input data.
Inspired by the principle of interpolation, we address this by inserting dense intermediate viewpoints to bridge gaps and enforce continuity across views.

To this end, we propose \textbf{Dehallu3D}, a novel framework for single-image 3D mesh generation that effectively eliminates hallucination-induced outliers.
The core design is a plug-and-play module termed Cyclic View Consistency Refinement (CVCR). Specifically, CVCR renders dense views beyond the orthographic views and enforces consistency across adjacent views to bridge the discontinuities between sparse views. This consistency supervision helps mitigate hallucinations by eliminating outliers. 
As increasing view density inevitably leads to higher computational cost, we count on a balanced angular interval to reach both reconstruction quality and efficiency. 
Moreover, to prevent over-smoothing during view consistency enforcement, we apply adaptive smoothness constraints to preserve sharp geometric features such as spikes.

To the best of our knowledge, we are the first to explicitly cope with the outliers in the context of 3D reconstruction. Since there are no available evaluation metrics to assess the geometry fidelity from this perspective, we design an Outlier Risk Measure (ORM) metric to assess the quality of reconstructed 3D contents, particularly from the viewpoint of outliers.
Overall, the contributions of this work are:
\begin{itemize}
    \item We propose \textbf{Dehallu3D}, a novel framework for high-fidelity 3D mesh generation from a single image that effectively mitigates hallucinations by eliminating outliers.
    \item We introduce a plug-and-play module \textbf{CVCR} that ensures the consistency of adjacent views and smoothness variations in the mesh, thereby preserving fine details while suppressing hallucinated outliers. 
    \item We devise the \textbf{Outlier Risk Measure (ORM)}, a new metric tailored to evaluate geometric fidelity.
    \item Our experiments show that Dehallu3D achieves superior performance, enabling high-quality single-image-to-3D generation.
\end{itemize}

%% file: sec/2_relatedwork.tex
\section{Related work}
\label{sec:formatting}
\subsection{Singe image-to-3D generation}
Due to the inherent lack of geometric information in 2D images, 3D generation from a single image is an ill-posed problem. DreamFusion \cite{poole2022dreamfusion} adopts SDS techniques to generate 3D assets. It distills 3D geometry and appearance from large and pre-trained image diffusion models \cite{saharia2022photorealistic}. Thus, it exploits the powerful 2D priors while avoiding the reliance on 3D assets which are less abundant than 2D image datasets. 
However, SDS-based methods \cite{shen2023make, liu2024sherpa3d, shen2023anything, xu2023neurallift, liu2026isotropic3d} may face the multi-face problems, also known as the ``Janus'' problem due to the lack of 3D priors. Zero123 \cite{liu2023zero} finetunes Stable Diffusion \cite{rombach2022high} to synthesis novel views on given camera poses and address this problem.
These Optimization-based methods are constrained by slow generation speeds, and the following works \cite{chen2024it3d, li2024focaldreamer} improve the results in speed. 
3D native generators \cite{hong2023lrm, xu2024instantmesh, wang2024pf, xu2023dmv3d, lu2024large, tochilkin2024triposr, xiang2025structured} directly generate 3D content from a single 2D image, bypassing multi-view supervision. While they avoid multi-view consistency issues, they rely heavily on the model’s ability to learn complex 3D structures from limited 2D input. This can lead to challenges in complex or uncommon scenarios, where the model may struggle to produce high-quality results and the generated 3D content may lack detail or accuracy.
In contrast, One-2-3-45 \cite{liu2023one2345} proposes to integrate 2D generative models with multi-view 3D reconstruction, thus achieving a better performance in both quality and efficiency. Many other works follow this paradigm and improve the results a lot, in either speed or quality \cite{peng2024charactergen, wang2024crm, li2024icontrol3d, wu2024unique3d, yu2025fancy123, tang2025cycle3d}. 
This paradigm has gained popularity due to its high reconstruction quality. However, it still faces a critical issue: large gaps and discontinuities between the multi-views.
This leads to a degradation in the quality of subsequent reconstruction.
Therefore, we propose to improve the reconstruction quality by bridging the gap between multi-view images.

\subsection{Mesh representation for reconstruction}
3D representations are significant for 3D reconstruction tasks, as they directly influence the efficiency and quality of reconstructed geometry and texture. 
Considerable progress has been made in various 3D representation techniques, such as Neural Radiance Fields (NeRF) \cite{mildenhall2021nerf, muller2022instant}, point clouds \cite{lionar2023nu}, and 3D Gaussian \cite{kerbl20233d, lu2024large}.
However, meshes continue to dominate as the preferred representation, owing to their well-established rendering pipeline. 
Some studies \cite{lorensen1998marching, yang2024hi3d, boss2025sf3d, wu2024unique3d, long2024wonder3d, yu2025fancy123} utilize techniques such as Structure from Motion (SfM), Multi-View Stereo (MVS), and mesh surface reconstruction for 3D modeling. 
Subsequently, differentiable meshes \cite{shen2021deep, shen2023flexible, wei2023neumanifold, tochilkin2024triposr} are proposed for 3D optimization tasks. A differentiable mesh is a hybrid 3D representation that combines both implicit and explicit surface representations, such as Signed Distance Functions (SDFs) and meshes. Although alternative 3D representations offer distinct benefits, they incur significant memory overhead. In contrast, our approach retains meshes as the 3D representation, leveraging their efficient rendering and compact storage properties.

%% file: sec/4_method.tex
\section{Method}
\subsection{Overview}
\begin{figure*}[!htb]
    \centering
    \includegraphics[width=0.99\textwidth]{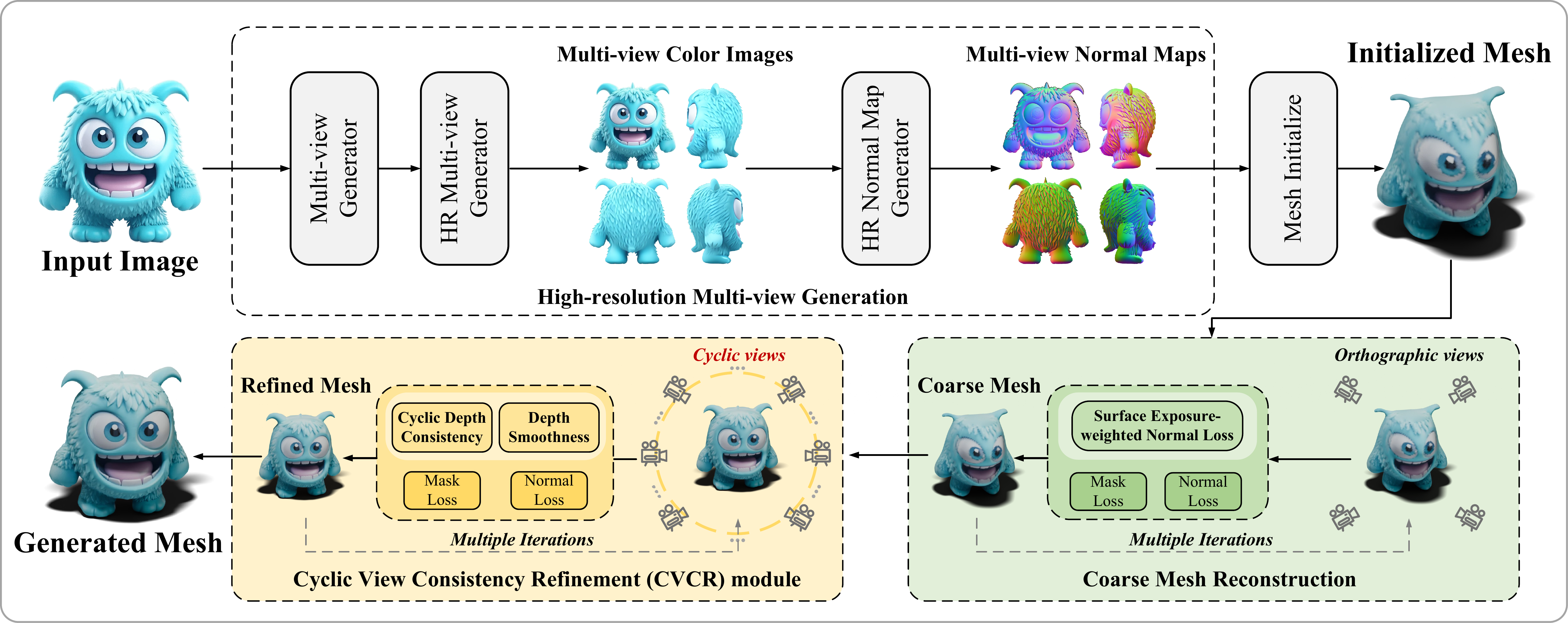}
    \caption{\textbf{Overview of Dehallu3D.} Dehallu3D first generate orthographic multi-view color images and corresponding normal maps, which are used to initialize a coarse mesh. Next, a globally plausible mesh is quickly constructed through the {Coarse Mesh Reconstruction} stage. Finally, the proposed \textbf{Cyclic View Consistency Refinement (CVCR)} module is employed to mitigate outliers and further refine the mesh.
    }
    \label{fig_pipeline}
\end{figure*}

In this section, we propose \textbf{Dehallu3D}, a novel single-image 3D mesh generation framework that produces high-fidelity meshes while effectively mitigating outliers during reconstruction. 
As illustrated in Figure~\ref{fig_pipeline}, the pipeline of Dehallu3D begins with high-resolution multi-view generation. 
Given an input image, we first generate four orthographic multi-view color images along with their corresponding normal maps. 
We then apply a fast initialization method to generate a rough mesh.
To progressively refine the mesh, Dehallu3D adopts a two-stage optimization pipeline:
(1) \textbf{Coarse Mesh Reconstruction.} This stage focuses on globally correcting the mesh topology using differentiable rendering. We introduce a surface exposure-weighted normal loss to prioritize geometry constraints from high-visibility regions, ensuring reliable global structure.
(2) \textbf{Cyclic View Consistency Refinement (CVCR).} To mitigate mesh outliers and recover fine details, we propose CVCR, a plug-and-play refinement module that enforces consistency across cyclically adjacent viewpoints and adaptive smoothness noise to prevent over-smoothing.
Through this coarse-to-fine strategy, Dehallu3D achieves robust and detailed mesh reconstruction from a single image.

\subsection{Mesh initialization}
The quality of mesh initialization critically influences subsequent optimization stages and final results. 
Therefore, an effective mesh initialization method should establish approximately correct topological structures.
To this end, following \cite{wu2024unique3d}, we adopt a fast initialization process to directly capture the complete topological connection features of the visible area through the front and back views.
Specifically, given an input image, we first use the existing high-resolution multi-view generators to obtain multi-view color images and their corresponding normal maps. 
Next, we utilize the normal maps from the front and back views to improve the reliability of depth estimation through integration and random rotations, and combine this with Poisson reconstruction technology to generate a high-quality initialized mesh. 

\subsection{Coarse mesh reconstruction}
To effectively improve the quality of the mesh and mitigate outliers caused by error accumulation due to multi-view inconsistencies during reconstruction, we adopt a two-stage optimization pipeline.
In the coarse reconstruction stage, rapidly correcting the overall shape of the mesh is crucial. We perform mesh reconstruction and optimization based on differentiable rendering. The loss function at the current stage is defined as
\begin{equation}
    \mathcal{L}_{coarse}=\mathcal{L}_{mask}+\mathcal{L}_{normal}+\mathcal{L}_{SE}.
\end{equation}

In $\mathcal{L}_{coarse}$, we propose a \textit{surface exposure-weighted normal loss} $\mathcal{L}_{{SE}}$ to effectively fuse normal information from multiple views, which is defined as
\begin{equation}
\begin{split}
    \mathcal{L}_{{SE}} &= \sum_{v \in \mathcal{V}} \sum_{i}^4 \epsilon_{i}^v \cdot | N_v^R - N_{i}^v |_2^2,\\
    \epsilon_{i}^v &= m_{i}^v \cdot \frac{A_{i}^v}{\sum_{j} m_{j}^v A_{j}^v}.
\end{split}
\end{equation}
Here, $\mathcal{V}$ denotes the set of mesh vertices, $N_v^R$ is the extracted surface normal of vertex $v$, and $N_{i}^v$ represents the reference normal of vertex $v$ in view $i$. 
The visibility mask $m_{i}^v \in {0, 1}$ indicates whether vertex $v$ is visible in view $i$. 
The projected surface area $A_{i}^v$ is the sum of the projected areas of the triangular faces associated with vertex $v$ in view $i$, reflecting the degree of observability of the vertex in that view. 
Views with larger projected areas typically correspond to core regions of the mesh, providing stronger geometric constraints that help quickly correct the coarse mesh's global structure.
The weighting term $\epsilon_{i}^v$ is derived from the projected area of vertex $v$ in view $i$, dynamically prioritizing views with higher projected areas while suppressing the influence of views with low visibility. This enables robust global shape optimization during the coarse reconstruction stage.


The mask-based loss is defined as
\begin{equation}
    \mathcal{L}_{mask}=\sum_{i=1}^4\|{M}_i-M_i^{R}\|_2^2.
\end{equation}
Here, ${M}_i$ and $M_i^{R}$ denote the alpha channel values of the generated color image under view $i$ and the corresponding rendered color image, respectively.

The normal-based loss is defined as
\begin{equation}
    \mathcal{L}_{normal}=\sum_{i=1}^4 \|{N}_i-N_i^{R}\|_2^2.
\end{equation}
Here, ${N}_i$ denotes the generated normal map under view $i$ and $N_i^{R}$ denotes the rendered normal map under view $i$.
$\mathcal{L}_{mask}$ and $\mathcal{L}_{normal}$ are both common MSE losses in 3D mesh reconstruction.
After the above optimization, the overall geometric structure of the mesh has been significantly improved. Next, we refine the mesh to further mitigate outliers during reconstruction.

\subsection{Cyclic view consistency refinement}
To further enhance geometric fidelity and mitigate outliers, we propose a \textbf{Cyclic View Consistency Refinement (CVCR)} module.
Unlike general orthographic view-consistent methods, CVCR explicitly models the cyclic relationship among adjacent views over a full 360$^\circ$ rotation, enforcing pairwise alignment between neighboring depth maps.
This module is designed to be \textit{plug-and-play}, making it easily integrable into a wide range of mesh reconstruction pipelines, regardless of the mesh initialization strategy.

The CVCR module optimizes the mesh using the following objective:
\begin{equation}
    \mathcal{L}_{{CVCR}} = \mathcal{L}_{{mask}} + \mathcal{L}_{{normal}} + \lambda_1 \mathcal{L}_{{DC}} + \lambda_2 \mathcal{L}_{{DS}}.
\end{equation}
Here, $\mathcal{L}_{{mask}}$ and $\mathcal{L}_{{normal}}$ are inherited from the coarse reconstruction stage.
The two newly introduced terms, $\mathcal{L}_{{DC}}$ and $\mathcal{L}_{{DS}}$, are described below.


Empirical evidence shows that depth maps of real-world objects maintain consistency across adjacent small-angle views when observed through a full 360$^\circ$ rotation. 
In contrast, abrupt depth discontinuities in such views frequently lead to obvious outliers in 3D meshes.
Building on this, we introduce a \textit{cyclic depth consistency loss} $\mathcal{L}_{{DC}}$ to encourage the reconstructed mesh to better adhere to real-world physical properties by enforcing consistency between depth maps rendered from adjacent viewpoints with small angular differences. 
This helps mitigate outliers in the mesh. The loss is defined as
\begin{equation}
\begin{split}
\mathcal{L}_{{DC}} &= \sum_{i=1}^V \left( 1 - \Delta(D_i^R, D_{i \bmod V + 1}^R) \right), \\
\Delta(D_i^R, D_j^R) &= \mathrm{SSIM}(D_i^R, D_j^R) \cdot \mathrm{CS}(D_i^R, D_j^R).
\end{split}
\end{equation}
Here, $V$ denotes the total number of views (we set $V=72$, yielding $360^{\circ} /72=5^{\circ}$ angular differences). $D_i^R$ is the rendered depth map from view $i$, and $D_{i \bmod V + 1}^R$ is its adjacent cyclic view. The similarity function $\Delta(\cdot)$ integrates the Structural Similarity Index Measure (SSIM) \cite{wang2004image} and Cosine Similarity (CS) to evaluate both structural and directional alignment. This design ensures robustness against pixel misalignment in adjacent depth maps caused by angular deviations between views.

While $\mathcal{L}_{{DC}}$ enforces global consistency, it may over-constrain regions with valid sharp features, leading to over-smoothing. 
To mitigate this, we introduce a complementary \textit{depth smoothness loss} $\mathcal{L}_{{DS}}$ that adaptively preserves sharp features in the mesh while mitigating hallucinated outliers based on image gradients, which is defined as
\begin{equation}
\begin{split}
\mathcal{L}_{{DS}} &= \sum_{i=1}^V \sum_{j,k}^{pixel} |\nabla D_i^{R(j,k)}| \cdot w_i^{j,k},\\
w_i^{j,k} &= \exp( -\|\nabla I_i^{R(j,k)}\|_2 ).\\
\end{split}
\end{equation}
Here, $I_i^R$ represents the rendered color image for view $i$, and $D_i^R$ denotes its corresponding depth map. 
The gradient magnitude of the depth map at pixel $(j, k)$ is defined as $|\nabla D_i^{R(j,k)}| = \sqrt{(\partial_x D_i^{R(j,k)})^2 + (\partial_y D_i^{R(j,k)})^2}$, capturing the intensity of depth changes. 
Similarly, the gradient magnitude of the color image at pixel $(j, k)$ is defined as $\|\nabla I_i^{R(j,k)}\|_2 = \sqrt{\|\partial_x I_i^{R(j,k)}\|_2^2 + \|\partial_y I_i^{R(j,k)}\|_2^2}$, reflecting color variations associated with geometric features.
The weight $w_i^{j,k}$ dynamically adjusts the strength of the smoothness constraint. 
In regions with high color image gradients where indicates potential sharp features, the weight reduces the smoothness penalty, preserving depth discontinuities. 

%% file: sec/5_exp.tex
\section{Experiment}

\subsection{Datasets and evaluation metrics}
Our experiments focus on both mesh appearance quality and mesh geometric quality.
Following previous work \cite{wu2024unique3d, wang2024crm, muller2022instant, yu2025fancy123}, we perform our experiments on the Google Scanned Objects (GSO) dataset \cite{downs2022google}. 
For a fair experimental comparison, all objects in the dataset are rendered at a resolution of $512 \times 512$ with Blender Cycles as input for all methods. 
All generated mesh results are normalized to the bounding box of $[-0.5,0.5]$ to ensure alignment.
For visual quality evaluation, we select elevation angles from $\{0,15,30\}$ and 8 evenly distributed azimuth angles to render each object generated from different methods, resulting in 24 views per object. 
We employ PSNR, SSIM, LPIPS, and Clip-Similarity (Clip-Sim) as metrics to evaluate the visual quality.
For geometric quality evaluation, we utilize Chamfer Distance (CD) and F-Score as metrics. More implementation details are demonstrated in the supplementary material. All experiments are conducted on a single NVIDIA GeForce RTX 4090 GPU.

\subsection{Outlier risk measure}
In 3D mesh reconstruction, mesh outliers impair geometric consistency and cause significant local errors. 
However, existing metrics often fail to capture the impact of these outliers.

To this end, we propose a novel evaluation metric based on Conditional Value-at-Risk (CVaR) \cite{li2024pointcvar} to measure the outlier degrees of the meshes. 
When computing CVaR, it is first necessary to determine the Value at Risk (VaR), that is, the maximum risk that may occur at a given probability level.
Given a set of risk values $\varphi=\{r_i\}_{i=1}^{\mathcal{R}}$ under a discrete risk distribution $\mathcal{D}$, the discrete CVaR is defined as 
\begin{equation}
    \begin{split}
        \mathrm{VaR_{\xi}(\varphi)}=\min\{r_i \in \varphi: \sum_{r_j \in \varphi}\mathbb{I}(r_i \geq r_j)\mathcal{D}(r_j) \geq \xi\},\\
        \mathrm{CVaR}_{\xi}(\varphi)=\frac{1}{1-\xi}\sum_{r_i \in \varphi}\mathbb{I}(r_i\geq \mathrm{VaR}_\xi(\varphi))\mathcal{D}(r_i)r_i.
    \end{split}
\end{equation}
Here, $\xi$ denotes the confidence level, $\mathbb{I}$ denotes the indicator function, and $\mathcal{D}$ is set to the empirical distribution $\mathcal{D}(r_i)=1/{\mathcal{R}}$.
Referring to the hypothesis in \cite{li2024pointcvar}, \textit{the presence of outliers in point clouds will increase the level of tail risk in its point risk distribution}, we convert the mesh into point clouds and measure the outlier degrees of different meshes by computing the tail risk in their point clouds.

In CVaR, $\varphi$ is used to quantify the risk of points in the point clouds.
Given a 3D mesh $\mathcal{M}$ and its point cloud is denoted as $\mathcal{P}=\{p_i\}_{i=1}^P$.
To evaluate this risk, we introduce an outlier scoring function $S(\mathcal{P})$.
Specifically, $S(\mathcal{P})$ integrates both global and local outlier measures in the point cloud.
The global measure, denoted as $S_g(\mathcal{P})$, is derived from the reconstruction loss of the VAE.
Meanwhile, the local measure, $S_l(\mathcal{P})$, is computed based on the neighborhood density of the internal points, which is defined as 
\begin{equation}
\begin{split}
    S_l(\mathcal{P})=\frac{1}{P}\sum_{i=1}^P  \frac{d_i^k}{\frac{1}{|\mathcal{N}_i|}\sum\limits_{p_j\in \mathcal{N}_i}d_j^k}.
\end{split}
\end{equation}
Here, $\mathcal{N}_i \subseteq \mathcal{P}$ denotes the nearest neighbors of $p_j \in \mathcal{P}$ and $d_j^k$ denotes the distance from $p_j$ to its $k$-th nearest neighbor. 
We compute the ratio of the average local density of a point and its neighbors to reflect the local outlier degree of the point in $\mathcal{P}$. 
The average outlier degrees of all points in $\mathcal{P}$ is taken as the local outlier degree $S_l(\mathcal{P})$ of the point cloud.
We use the $k$-th distance to reflect the density of the point. The larger the $k$-th distance, the smaller the density of the point.
The lower the local density of the point $p_i$, the more likely it is to be an outlier. 

The final outlier scoring function is $S(\mathcal{P})=S_l(\mathcal{P}) + \lambda S_g(\mathcal{P})$. 
We measure the risk of points $\phi$ by $S(\mathcal{P})$ and then compute the tail risk in $\phi$ as the final outlier score of $\mathcal{M}$. 
We refer to the outlier score obtained through the above process as the \textbf{Outlier Risk Measure ({ORM})}.
The greater the ORM, the more outliers the meshes have. 
We hope to generate meshes with low ORM values, that is, meshes with relatively few outliers.

\begin{table*}[!htb]
    \centering
    \begin{tabular}{l|c|c|c|c|c|c}
    \bottomrule[1.25pt]
      & \multicolumn{4}{c|}{Appearance metrics} & \multicolumn{2}{c}{Geometry metrics} \\
     \hline
     Methods & PSNR$\uparrow$ & SSIM$\uparrow$ & LPIPS$\downarrow$ & Clip-Sim$\uparrow$ & CD$\downarrow$ & F-Score$\uparrow$ \\
    \hline
    Wonder3D \cite{long2024wonder3d}   & 20.4963          & {0.8908}          & 0.1851          & 0.6970          & 0.02183          & 0.3580          \\
    TripoSR \cite{tochilkin2024triposr}    & 20.5309          & 0.8897          & 0.1841          & 0.7146          & 0.02241          & 0.3847          \\
    InstantMesh \cite{xu2024instantmesh} & 20.8954          & {0.8903}          & 0.1749          & \underline{0.7538}          & 0.02198          & {0.4046}          \\
    CRM   \cite{wang2024crm}     & {21.1265}          & 0.8889          & {0.1720}          & 0.7191          & {0.02163}          & 0.3967          \\
    Unique3D  \cite{wu2024unique3d}  & {20.9795}          & 0.8882          & {0.1742}          & {0.7493}          & {0.02175}          & \underline{0.4073}    \\
    SF3D    \cite{boss2025sf3d}    & \underline{21.3257}          & \underline{0.8912}          & \underline{0.1537}          & 0.7463          & \underline{0.02144}          & 0.3765        \\
    \textbf{Dehallu3D (Ours)}    & \textbf{21.8407} & \textbf{0.8966} & \textbf{0.1453} & \textbf{0.7753} & \textbf{0.02023} & \textbf{0.4212} \\
    \toprule[1.25pt]
  \end{tabular}
  \caption{Quantitative comparison results. We mark the best scores in \textbf{bold} and the second-best scores with an \underline{underline}.}
  \label{tab_dinliang}
\end{table*}

\begin{figure*}[!htb]
    \centering
    \includegraphics[width=1\textwidth]{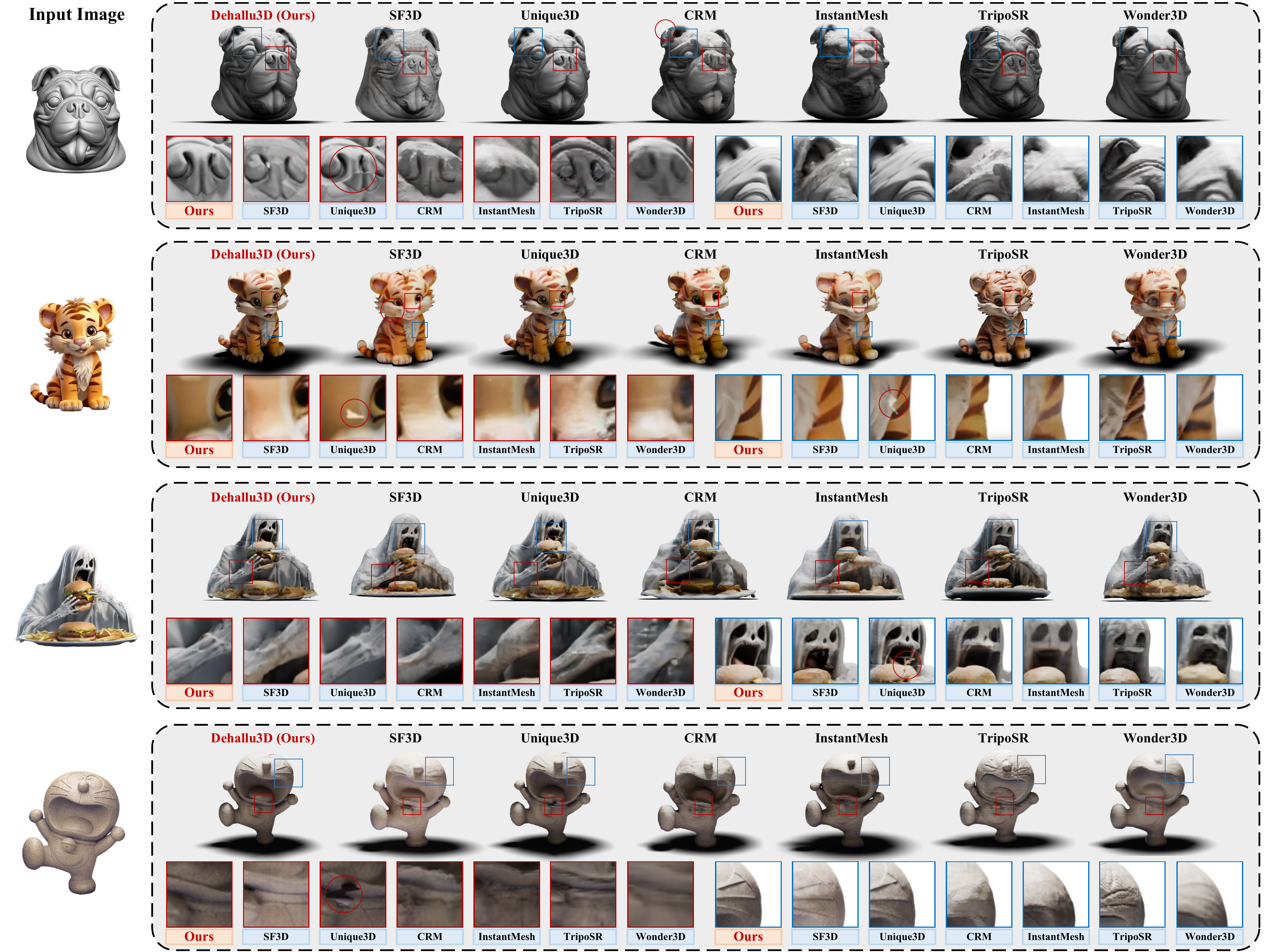}
    \caption{Qualitative comparison of different methods in mesh reconstruction. The red and blue boxes mark specific regions in the meshes of other methods for comparison with the corresponding optimized regions in the mesh generated by Dehallu3D, while red circles highlight defects in meshes.}
    \label{fig_dingxing}
\end{figure*}

\begin{figure*}[!htb]
    \centering
    \includegraphics[width=0.99\textwidth]{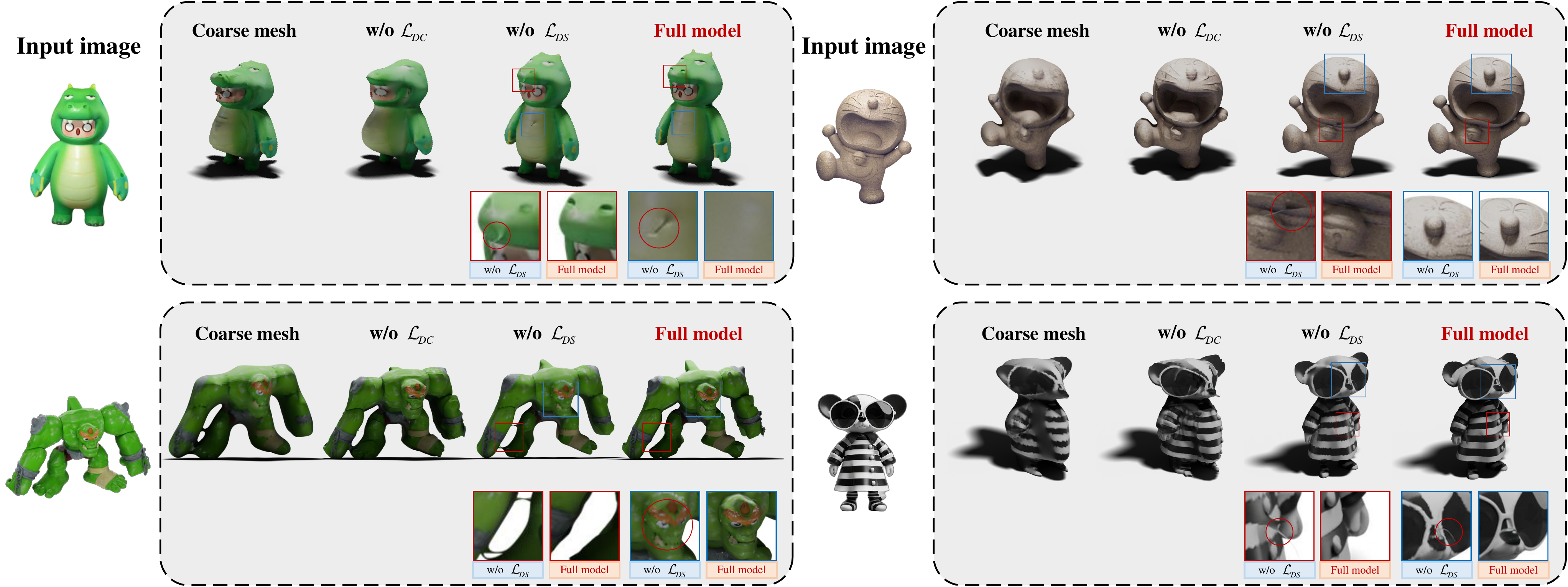}
    \caption{Qualitative comparison results for ablation study on $\mathcal{L}_{DC}$ and $\mathcal{L}_{DS}$ in CVCR module.}
    \label{fig_ablation}
\end{figure*}

\begin{table*}[!htb]
    \centering
    \begin{tabular}{ccc|c|c|c|c|c|c}
    \bottomrule[1.25pt]
     \multicolumn{3}{c|}{} & \multicolumn{4}{c|}{Appearance metrics} & \multicolumn{2}{c}{Geometry metrics} \\
     \hline
    $\mathcal{L}_{SE}$ & $\mathcal{L}_{DC}$ & $\mathcal{L}_{DS}$ & PSNR$\uparrow$         & SSIM$\uparrow$       & LPIPS$\downarrow$    & Clip-Sim$\uparrow$      & CD$\downarrow$      & F-Score$\uparrow$       \\
    \hline
    \color{red}{\ding{55}} & \color{red}{\ding{55}} & \color{red}{\ding{55}} & 16.8972 & 0.8474 & 0.2283 & 0.6315 & 0.02572 & 0.3201 \\
    \color{teal}{\ding{52}} & \color{red}{\ding{55}} & \color{red}{\ding{55}} & 17.6534 & 0.8615 & 0.2102 & 0.6678 & 0.02449 & 0.3468 \\
    \color{red}{\ding{55}} & \color{teal}{\ding{52}} & \color{red}{\ding{55}} & 20.9276 & 0.8823 & 0.1697 & 0.7392 & 0.02217 & 0.3879 \\
    \color{red}{\ding{55}} & \color{red}{\ding{55}} & \color{teal}{\ding{52}} & 18.3124 & 0.8749 & 0.2019 & 0.6834 & 0.02332 & 0.3545 \\
    \color{red}{\ding{55}} & \color{teal}{\ding{52}} & \color{teal}{\ding{52}} & \underline{21.1973} & \underline{0.8894} & \underline{0.1531} & \underline{0.7403} & \underline{0.02114} & \underline{0.4097} \\
    \color{teal}{\ding{52}} & \color{teal}{\ding{52}} & \color{teal}{\ding{52}} & \textbf{21.8407} & \textbf{0.8966} & \textbf{0.1453} & \textbf{0.7753} & \textbf{0.02023} & \textbf{0.4212} \\
    \toprule[1.25pt]
    \end{tabular}
    \caption{Quantitative comparison results for ablation study on the proposed losses.}
    \label{tab_ablation}
\end{table*}

\begin{figure}[!htb]
    \centering
    \includegraphics[width=1\linewidth]{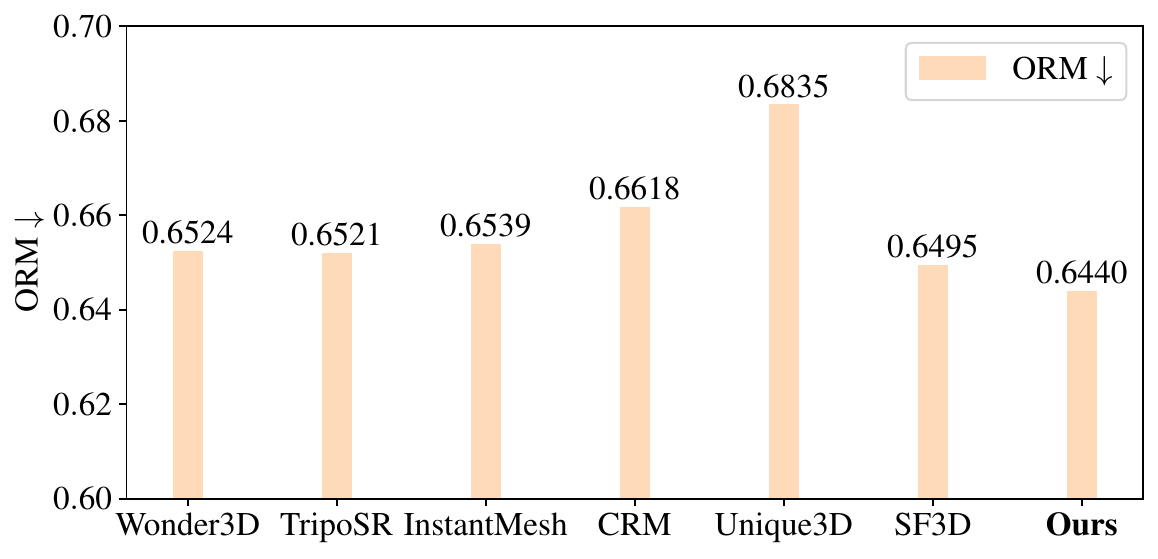}
    \caption{Comparison of ORM results across all methods.}
    \label{fig_ORM}
\end{figure}

\begin{table*}[!htb]
\centering
\begin{tabular}{l|c|c|c|c|c|c|c}
\bottomrule[1.25pt]
\multicolumn{1}{c|}{} & \multicolumn{4}{c|}{Appearance metrics} & \multicolumn{2}{c|}{Geometry metrics} & Time \\
\hline
{Angles} & PSNR$\uparrow$         & SSIM$\uparrow$       & LPIPS$\downarrow$    & Clip-Sim$\uparrow$      & CD$\downarrow$      & F-Score$\uparrow$ & seconds$\downarrow$ \\
\hline
CVCR (3°)  & \textbf{21.8628} & \textbf{0.8971} & \underline{0.1455} & \textbf{0.7764} & \underline{0.02030} & \underline{0.4203} & 208.1 \\
CVCR (5°)  & \underline{21.8407} & \underline{0.8966} & \textbf{0.1453} & \underline{0.7753} & \textbf{0.02023} & \textbf{0.4212} & 163.3 \\
CVCR (10°) & 21.3687 & 0.8879 & 0.1527 & 0.7591 & 0.02134 & 0.4118 & \underline{129.5} \\
CVCR (15°) & 20.8415 & 0.8763 & 0.1671 & 0.7459 & 0.02221 & 0.3936 & \textbf{110.8} \\
\toprule[1.25pt]
\end{tabular}
\caption{Ablation study of angular intervals between adjacent views in the CVCR module.}
\label{tab_cvcr}
\end{table*}

\subsection{Comparisons}
\subsubsection{Quantitative comparison} 
To validate the superiority of our method, we conduct comprehensive comparisons with SOTA methods.
Considering the 3D representation method and open-source availability, we select SF3D \cite{boss2025sf3d}, Unique3D \cite{wu2024unique3d}, CRM \cite{wang2024crm}, InstantMesh \cite{xu2024instantmesh}, TripoSR \cite{tochilkin2024triposr}, Wonder3D \cite{long2024wonder3d} as comparison methods in the experiments.
The quantitative comparison results are shown in Table~\ref{tab_dinliang}.
Specifically, {Dehallu3D} achieves the best performance across multiple metrics for both visual quality and geometric quality.
The best performance of our method on all metrics highlights Dehallu3D, which exploits depth consistency and depth smoothness to boost the quality of the generated 3D meshes.

We also compare the ORM results across all methods, as shown in Figure~\ref{fig_ORM}. Our method achieves the lowest ORM value, while Unique3D exhibits the highest, consistent with its performance in both quantitative and qualitative experiments. We conducted experiments on the ORM-human perception correlation in the supplementary materials.


\subsubsection{Qualitative comparison} 
In this section, we visually evaluate the performance of the proposed method and its effectiveness in 3D mesh generation tasks. 
The primary goal of the qualitative experiments is to demonstrate the model's capability in handling mesh outliers and to highlight its strengths in maintaining geometric consistency and preserving fine details through visual results.
We select a set of example samples from existing studies. 
A comparative visualization of the meshes reconstructed by our method against other methods is presented in Figure~\ref{fig_dingxing}.
SF3D generally achieves desirable 3D mesh generation from input images, but the texture of the entire mesh remains relatively blurry.
Compared to SF3D, Dehallu3D excels in capturing texture details in the mesh, enabling better reconstruction of detailed features in the input images.

Next, we further compare the meshes generated by Dehallu3D and Unique3D.
While Unique3D demonstrates the ability to reconstruct texture details, it fails to consider the overall depth consistency and depth smoothness of the mesh, making it susceptible to mesh outliers during reconstruction. This issue in turn causes a decrease in the overall quality of the mesh.
The remaining methods including CRM, InstantMesh, TripoSR, and Wonder3D, generate meshes that are inferior in quality compared to our method. The reconstruction results of these methods lack fine texture details and precise geometry.
More qualitative experimental results are presented in the supplementary material.

\subsection{Ablation study}
The quantitative results in Table~\ref{tab_ablation} demonstrate the synergistic impact of each proposed loss. 
Using only $ \mathcal{L}_{SE} $ fails to mitigate mesh outliers. Introducing $ \mathcal{L}_{DC} $ reduces outliers but can lead to over-smoothing of depth details due to the lack of $\mathcal{L}_{DS}$. Combining $ \mathcal{L}_{DS} $ and $ \mathcal{L}_{DC} $ significantly enhances mesh quality. However, omitting $ \mathcal{L}_{SE} $ results in a poor-quality coarse mesh, limiting the effectiveness of subsequent optimization.

As shown in Figure~\ref{fig_ablation}, we further demonstrate the synergistic effect of $\mathcal{L}_{DC}$ and $\mathcal{L}_{DS}$ through qualitative comparisons, highlighting their contributions to reducing mesh outliers and enhancing mesh quality. 
Without $\mathcal{L}_{DS}$, the generated mesh exhibits noticeable protrusions and surface irregularities. 
Without $\mathcal{L}_{DC}$, the mesh quality is significantly degraded due to the absence of cross-view consistency constraints. 
Combining $\mathcal{L}_{DC}$ and $\mathcal{L}_{DS}$ enables synergistic optimization, substantially improving mesh quality.


\begin{figure}[t]
    \centering
    \includegraphics[width=1\linewidth]{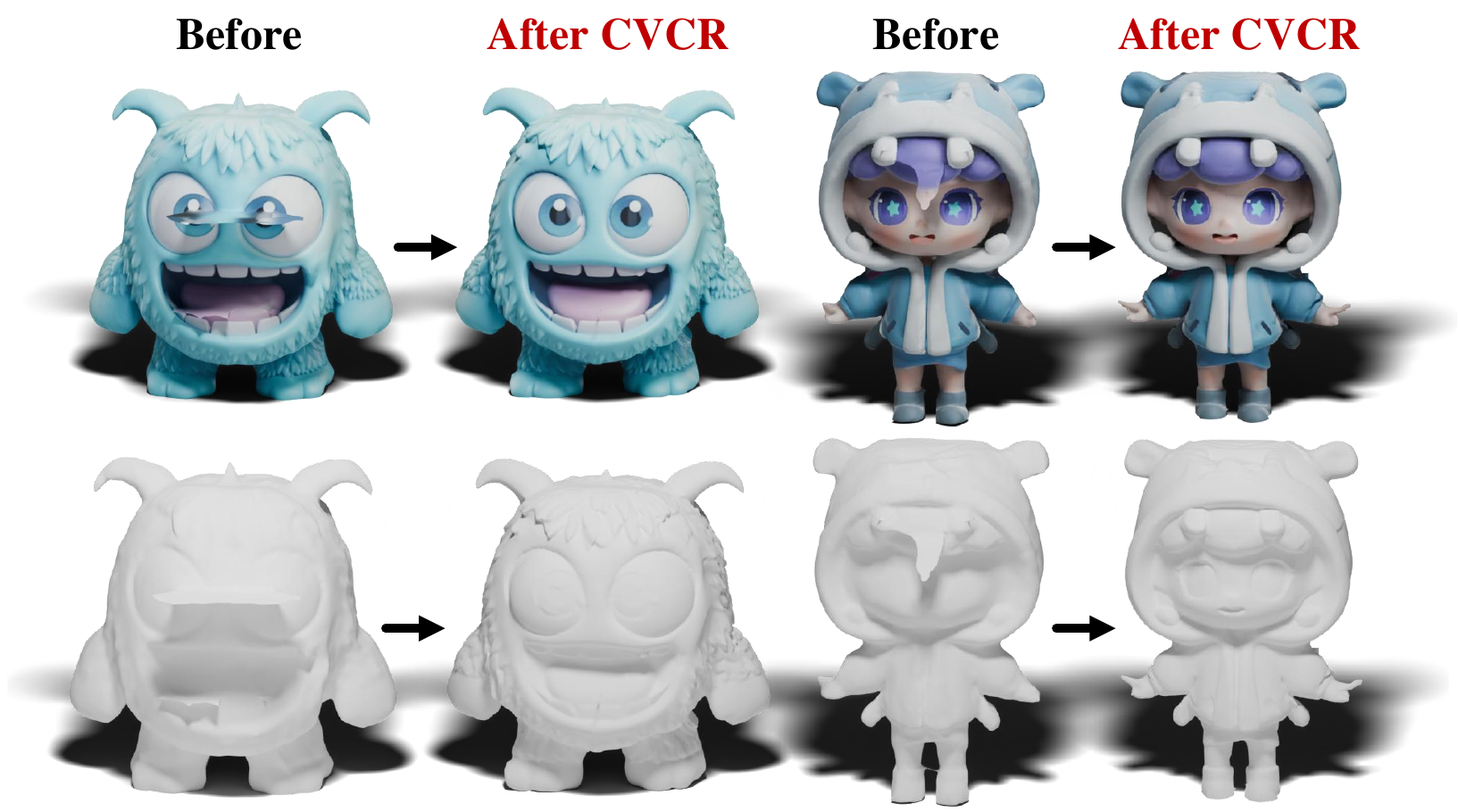}
    \caption{After applying the CVCR module to Unique3D.}
    \label{fig_improve}
\end{figure}

As presented in Table~\ref{tab_cvcr}, we analyze the impact of the angular interval for the dense views in our CVCR module. 
While larger intervals (e.g., $10^{\circ}$ or $15^{\circ}$) degrade performance due to sparse supervision, a very dense $3^{\circ}$ interval also leads to flawed results.
Specifically, the $3^{\circ}$ interval shows slightly worse geometric fidelity (CD, F-Score) and perceptual quality (LPIPS) compared to the $5^{\circ}$ interval.
We attribute this to potential {over-smoothing}; at such a small gap, the cyclic depth consistency loss ($\mathcal{L}_{DC}$) becomes overly strict, penalizing valid sharp geometric features by mistaking them for discontinuities. 
This effect appears to overpower the protection from our adaptive smoothness loss ($\mathcal{L}_{DS}$).
Furthermore, a smaller angular interval directly increases the number of required rendering views, consequently leading to a higher time overhead. Critically, this increased time expenditure does not proportionally guarantee superior results, a trend consistent with our prior analysis. Therefore, prioritizing a trade-off between quality and efficiency, we adopted a balanced angular interval of $\mathbf{5^{\circ}}$, which offers an acceptable computational overhead and delivers satisfactory performance.



Finally, we integrate the proposed plug-and-play CVCR module into the Unique3D method, which suffers from noticeable mesh outliers caused by multi-view inconsistencies.
As shown in Figure~\ref{fig_improve}, the quality of the meshes in Unique3D significantly improved after incorporating the CVCR module.

%% file: sec/6_conclusion.tex
\section{Limitations}
To achieve high fidelity, Dehallu3D introduces dense view rendering within its CVCR module, which inevitably leads to additional time overhead. This is a deliberate trade-off that prioritizes geometric accuracy over inference speed, making our method particularly suitable for accuracy-critical applications, such as 3D printing.

Nevertheless, we recognize the importance of efficiency in certain contexts. In the future, we plan to focus on improving reconstruction efficiency. Our efforts will target the optimization of this dense refinement stage, for instance, by enhancing parallel processing or developing more efficient algorithms for view rendering and comparison.



\section{Conclusion}
In this paper, we present \textbf{Dehallu3D} for single-image 3D reconstruction that mitigates model hallucinations by eliminating outliers by enforcing adjacent view consistency. The CVCR module is designed to be plug-and-play, rendering dense viewpoints to bridge gaps across sparse views, a common issue in multi-view generation methods. In addition, adaptive smoothness is applied to prevent over-smoothing caused by enforcing view consistency. To quantify geometric fidelity, we introduce a novel metric specifically designed to evaluate 3D quality by detecting outliers in meshes. Extensive experiments demonstrate that our method achieves the SOTA performance in both visual quality and geometric accuracy, significantly reducing the occurrence of outliers. And we validated that the proposed metric ORM aligns well with human perception in terms of hallucinated outlier assessment.

%% file: sec/acknowledgement.tex
\section*{Acknowledgment}
This work is supported by the National Natural Science Foundation of China, Young Scientists Fund (C Class)(No.62506251) and the National Natural Science Foundation of China (No.61773270, No.62176170).